\documentclass[10pt,twocolumn,letterpaper]{article}

\usepackage{cvpr}
\usepackage{times}
\usepackage{epsfig}
\usepackage{graphicx}
\usepackage{amsmath}
\usepackage{amssymb}
\usepackage{amsmath}
\usepackage{algorithm}
\usepackage[noend]{algpseudocode}
\usepackage{breqn}

\usepackage[breaklinks=true,bookmarks=false]{hyperref}

\cvprfinalcopy % *** Uncomment this line for the final submission

\def\cvprPaperID{****} % *** Enter the CVPR Paper ID here

\begin{document}

%%%%%%%%% TITLE
\title{Deep Siamese Networks with Bayesian non-Parametrics for Video Object Tracking}

\author{Anthony D. Rhodes \\
 Department of Mathematics\\
 Portland State University\\
Portland, OR 97207\\
 {\tt\small arhodes@pdx.edu}
% For a paper whose authors are all at the same institution,
% omit the following lines up until the closing ``}''.
% Additional authors and addresses can be added with ``\and'',
% just like the second author.
% To save space, use either the email address or home page, not both
\and
Manan Goel\\
Intel Corporation\\
{\tt\small}
}

\maketitle
%\thispagestyle{empty}

%%%%%%%%% ABSTRACT
\begin{abstract}
We present a novel algorithm utilizing a deep Siamese neural network as a general object similarity function in combination with a Bayesian optimization (BO) framework to encode spatio-temporal information for efficient object tracking in video. In particular, we treat the video tracking problem as a dynamic (i.e. temporally-evolving) optimization problem. Using Gaussian Process priors, we model a dynamic objective function representing the location of a tracked object in each frame. By exploiting temporal correlations, the proposed method queries the search space in a statistically principled and efficient way, offering several benefits over current state of the art video tracking methods.
\end{abstract}
\section{Introduction}

The problem of tracking an arbitrary object in video, where an object is identified by a single bounding-box in the first frame, requires both a robust similarity function and an efficient method for querying plausible locations of the object in subsequent frames.  Early video tracking approaches have included feature-based approaches and template matching algorithms [1] that attempt to track specific features of an object or even the object as a whole.
Feature-based approaches use local features, including points and edges, keypoints [2], SIFT features [3], HOG features [4] and deformable parts [5]. Conversely, template-based methods take the object as a whole – offering the potential advantage that they treat complex templates or patterns that cannot be modeled by local features alone. \\ \indent Through the course of a video, an object can potentially undergo a variety of different  visual transformations, including rotation, occlusion, changes in scale, illumination changes, etc., that pose significant challenges for tracking. In order to obtain a robust template matching for video tracking, researchers have developed a host of methods, including mean-shift [6] and cross-correlation filtering which entails convolving a template over a search region; significant advances to cross-correlation filtering for video tracking include MOSSE [7] adaptive correlation filter and the MUSTer algorithm [8] which draws influence from cognitive psychology in the design of a flexible object representation using long and short-term memory stored by means of an integrated correlation filter. 
\\ \indent More recently, deep learning models have been applied to video tracking to leverage the benefits of learning complex functions from large data sets.  While deep models offer the potential of improved robustness for tracking, they have nevertheless presented two significant challenges to tracking research to date. First, many deep tracking models are too slow for practical use due to the fact that they require online training, and, second, many deep trackers, when trained offline, are based on classification approaches, so that they are limited to class-specific searches and frequently require the aggregation of many image patches (and thus many passes through the network) in order to locate the object [9]. In light of these difficulties, several contemporary state of the art deep learning-based tracking models have been developed as generic object trackers in an effort to obviate the need for online training and to also improve the generalizability of the tracker. [10] applies a regression-based approach to train a generic tracker, GOTURN, offline to learn a generic relationship between appearance and motion; several deep techniques additionally incorporate motion and occlusion models, including particle filtering methods [11] and optical flow [12]. 
\\ \indent [13] demonstrated the power of deep Siamese networks (see section 2.1) based on [14], achieving a new state of the art for generic object matching for video tracking. Remarkably, the SINT algorithm delivered state of the art performance despite the fact that it was not equipped with any model updating, no occlusion detection, and no explicit geometric or feature matching components. [15,19] extended this work to achieve state of the art Siamese-based tracking while operating at frame rates beyond real-time by exploiting a fully-convolutional network structure. 
Even with these recent successes in video object tracking, there nevertheless exists a void in state of the art video tracking workflows that fully integrate deep learning models with classical statistics and machine learning approaches. Most state of the art video trackers lack – for instance – a capacity to generate systematic “belief” states (e.g. through explicit error and uncertainty measures), or ways to seamlessly incorporate contextual and scene structure, or to adaptively encode temporal information (e.g. by imposing intelligent search stopping conditions and bounds) and the ability to otherwise directly and inferentially control region proposal generation or sampling methods in a precise and principled way. To this end, we believe that the fusion of deep models with classical approaches can provide a necessary incubation for “intelligent” computer vision systems capable of high-level vision tasks in the future (e.g. scene and behavior understanding). 
In the current work we present the first integrated dynamic Bayesian optimization framework in conjunction with deep learning for object tracking in video. 

\section{Siamese Networks}

We adopt the Siamese network-based approach for one-shot image recognition from [15] to learn a generic, deep similarity function for object tracking. The network learns a function $f(z,x)$ that compares an exemplar crop \textit{z} to a candidate crop \textit{x} and returns a high score if the two images depict the same object and a low score otherwise. For computer vision tasks, a natural candidate for the similarity function $f$ is a deep conv-net [16,17]. Following [14,15], a Siamese network applies an identical transformation \(\phi\) to both 
input image crops and then combines their representations using another function $g$ that is trained to learn a general similarity function on the deep conv-net features, so that \(f(z,x)=g(\phi(z),\phi(x))\).

The network is trained on positive and negative pairs, using logistic loss: 
\begin{equation}
    l(y,v)=\log(1+\exp(-yv)) 
  \end{equation} where \textit{v} is the real-valued score of an exemplar-candidate pair and $y  \in \{-1,+1\} $ is its ground-truth label. The parameters of the conv-net \(\theta\) are obtained by applying Stochastic Gradient Descent (SGD) to:

\begin{equation}
  \operatorname*{arg\,min}_\theta 
   \mathbb{E}_{(z,x,y)}[ L(y,f(z,x;\theta))]
  \end{equation} 
where the expecation in eq. (2) is computed over the data distribution. 

Pairs of image crops were obtained using annotated videos from the 2015 edition of ImageNet for Large Scale Visual Recognition Challenge [18] (ILSVRC); images were extracted from two different frames, at most a distance of \textit{T} frames apart; positive image exemplars were defined as a function of their center offset distance from the ground-truth and the network stride length. Image sizes were normalized for consistency during training [15]. 

We use a five-layer conv-net architecture [19], with pooling layers after the first and second layers, and stride lengths of 2 and 1 throughout. The final network output is a 22x22x128 tensor, as shown in Figure 1. 

\begin{figure}
 \begin{center}
  \includegraphics[width=0.7\linewidth]{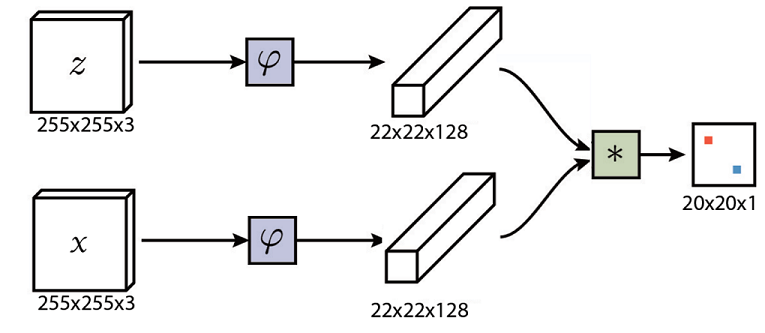}
  \end{center}
  \caption{The Siamese network \(\phi\) takes the exemplar image $z$ and search image $x$ as inputs. We then convolve (denoted by *) the output tensors to generate a similarity score. Similarity scores for a batch of sample search images are later rendered in a 20x20x1 search grid using a Gaussian Process (see section  3.3 for details). All images best viewed in color.}
  \label{fig:siamese}
\end{figure}

\section{Dynamic Bayesian Optimization}

[20] define object tracking in video as a dynamic optimization problem (DOP):
%\begin{equation}
\small
\begin{equation}
  DOP = \{\max\:f(\boldsymbol{x},t)\: s.t. \: x \in F(t) \subseteq \mathcal{S}, t \in \mathcal{T}\}
\end{equation}
where $\mathcal{S} \in \mathbb{R}^D$, with $\mathcal{S}$ in the search space; $f:\mathcal{S} \times \mathcal{T} \to\ \mathbb{R}$ is the temporally-evolving objective function which yields a maximum when the input $\boldsymbol{x}$ matches the ground-truth of the target object; $F(t)$ is the set of all feasible solutions $\boldsymbol{x} \in F(t) \subseteq \mathcal{S}$ at time \textit{t}.

Bayesian optimization is a sequential framework for optimizing an unknown, noisy and/or expensive objective function $f(\boldsymbol{x},t)$. BO works in two key stages: first, we generate a surrogate model to learn a latent objective function from collected samples; next, we determine plausible points to sample from the objective function in the search space. In the present work we use Gaussian Process Regression (GPR) to render the surrogate model. The second phase involves a secondary optimization of a surrogate-dependent \textit{acquisition function} $a(\boldsymbol{x},t)$, which strikes a balance between exploring new regions in the search space and exploiting information obtained from previous samples of the objective function. Common choices of acquisition functions include \textit{expected-improvement} (EI) and \textit{probability of improvement} (PI) functions [21]. We devise a novel acquisition function, which we call \textit{memory-score expected-improvement} (MS-EI), that demonstrated superior performance to EI and PI on our experimental data. We define MS-EI as:\small \begin{equation}
MS{\text -}EI(\boldsymbol{x})=
    \mu(\boldsymbol{x}))-f(\boldsymbol{x*})-\xi)\Phi(Z)+\sigma(\boldsymbol{x})\rho(Z)         
\end{equation}
where  $Z = \frac{\mu(\boldsymbol{x})-f(\boldsymbol{x}*)-\xi}{\sigma(\boldsymbol{x})},\boldsymbol{x}*=$ argmax $f(\boldsymbol{x}), \Phi$ and  $\rho $ denote the PDF and CDF of the standard normal distribution respectively [30,31]. We define $\xi=(\alpha \cdot$mean$[{f(x)]}_{D}\cdot{n^q)^{-1}}$; where $\alpha$ and $q$ are tunable parameters that depend on the scale of the objective function (we use $\alpha=1, q=1.1$); $D$ denotes the sample data set, and $n$ is the sample iteration number, with $|D|=n$; mean$[{f(x)]}_{D}$ is the sample mean of the previously observed values. Here $\xi$ serves to balance the exploration-exploitation trade-off to the specificity of a particular search. In this way, MS-EI employs a cooling schedule so that exploration is encouraged early in the search; however, the degree of exploration is conversely dynamically attenuated for exploitation as the search generates sample points with larger output values.

\subsection{Gaussian Processes}

A Gaussian Process (GP) defines a prior distribution over functions with a joint Normality assumption. We denote ${\hat{f}}$, the realization of the Gaussian process: $\hat{f}\sim\ \mathcal{GP}(\mu, K)$. Here the GP is fully specified by the mean $\mu: \mathcal{X}\to   
\mathbb{R}$ and covariance $K: \mathcal{X} \times \mathcal{X} \to\ \mathbb{R},   K((\boldsymbol{x},t),(\boldsymbol{x\textprime},t\textprime))=\mathbb{E}[(\hat{f}(\boldsymbol{x},t)-\mu(\boldsymbol{x},t))(\hat{f}(\boldsymbol{x\textprime},t\textprime)-\mu(\boldsymbol{x\textprime},t\textprime))],$ where $K(\cdot,\cdot) \leq 1$ and $\mathcal{X} = \mathcal{S}\times \mathcal{T}$. See  [21] for further details.

\subsection{Dynamic Gaussian Processes}

Following [22] we model a DOP $f(x,t)$ as a spatio-temporal GP where the objective function at time \textit{t} represents a slice of \textit{f} constrained at \textit{t}. This dynamic GP model will therefore encapsulate statistical correlations in space and time; furthermore the GP can enable tracking the location of an object, expressed as the temporally-evolving maximum of the objective function $f(x,t)$. 

Let $\hat{f}(\boldsymbol{x},t)\sim\ \mathcal{GP}(0,K(\{ \boldsymbol{x},t\},\{\boldsymbol{x\textprime},t\textprime\}))$, where $(\boldsymbol{x},t)\in\ \mathbb{R}^3$ ($\boldsymbol{x}$ is the bounding-box spatial location), and $K$ is the covariance function of the zero-mean spatio-temporal GP. For simplicity, we assume that $K$ is both stationary and separable of the form [22]:

\begin{equation}
  K(\hat{f}(\boldsymbol{x},t),\hat{f}(\boldsymbol{x\textprime},t\textprime))=K_{\mathcal{S}}(\boldsymbol{x},\boldsymbol{x\textprime}) \cdot K_{T}(t,t\textprime)
  \end{equation} where $K_{s}$ and $K_{T}$ are the spatio and temporal covariance functions, respectively. We use Mate\'rn kernel functions [21] in experiments and train the spatial and temporal covariance functions independently, following our separable assumption. 
  
  \begin{figure}
  \begin{center}
  \includegraphics[width=0.5\linewidth]{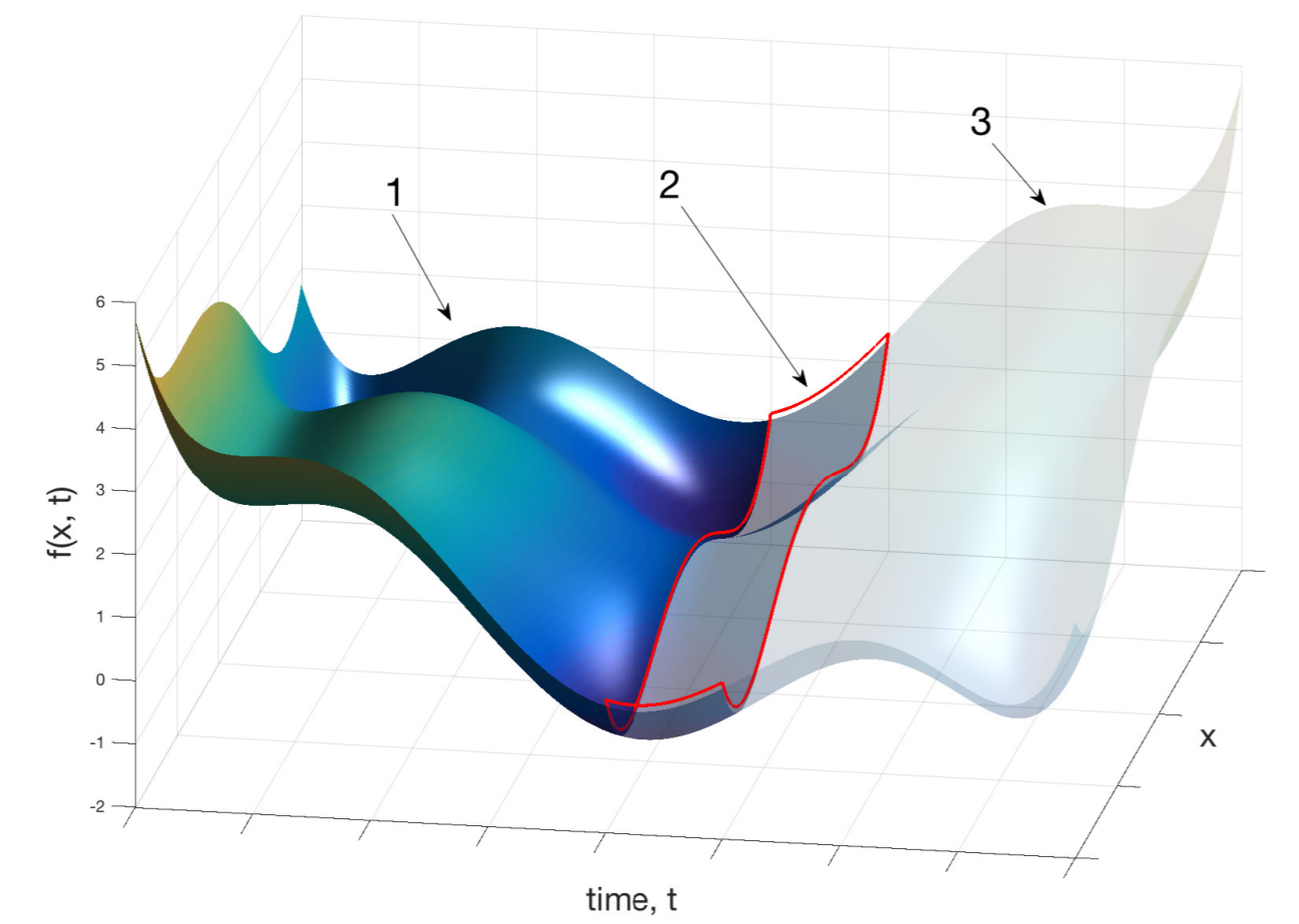}
  \end{center}
  \caption{Illustration of $\hat{f}(\boldsymbol{x},t)$ for DOP: Region (1) shows previous sample instances for time instances prior to time \textit{t}; region (2) depicts the bounded region of the search at time \textit{t}; region (3) represents future time slices. Image credit: [25].}
  \label{fig:DBO}
\end{figure}

\subsection{Siamese-Dynamic Bayesian Tracking Algorithm}

We now present the details of our \textit{Siamese-Dynamic Bayesian Tracking Algorithm} (SDBTA). The algorithm makes use of the previously-described deep Siamese conv-net. In the first step, we train the dynamic GP model. Then, for each “current” frame \textit{t} in the video containing \textit{T} total frames (consider $t=0$ the initial frame containing the ground-truth bounding-box for the target object), we render the GPR approximation over a resized search grid of size $d \times\ d$ (we use $d=20$ for computational efficiency), and then subsequently apply upscaling (e.g. cubic interpolation) over the original search space dimensions. In order to allow our algorithm to handle changes in the scale of the target object, each evaluation of an image crop is rendered by the Siamese network as a triplet score, where we compute the similarity score for the current crop compared to the exemplar at three scales: $ \{ 1.00-p, 1.00, 1.00 + p\} $, where we heuristically set $p=0.05$. The remaining algorithm steps are straightforward and detailed below. 

\begin{algorithm}
\caption{Siamese-Dynamic Bayesian Tracking Algorithm}\label{euclid}
\begin{algorithmic}
\State Train Dynamic GP model
\For i = 1,2,...T frames \textbf{do}
\For j = 1,2,...\{Max iterations per frame\} \textbf{do}
\State Calculate $\{\boldsymbol{x}_{i}, t_{i}\} = \text{arg max}_{x,t}$ 
$MS{\text -}EI(\boldsymbol{x},t)$
\State Query Siamese network $y_{i} \gets f(\boldsymbol{x}_{i}, t_{i})$
\State Augment new point to the data
\State Render GPR with set $\{y\}$ over $d \times\ d$ grid
\State Upsample grid data to dim. of search space $S$
\State Update current location of optimum over $S$
\EndFor
\EndFor
\State \indent \textbf{end for}
\State \textbf{end for}

\end{algorithmic}
\end{algorithm}
\section{Experimental Results}

We tested our algorithm using a subset of the VOT14 [32] and VOT16 [33] data sets, the "CFNET" video tracking data set [19], against three baseline video tracking models: template matching using normalized cross correlation (TM) [29] the MOSSE tracker algorithm [7], and ADNET (2017, CVPR), a state of the art, deep reinforcement learning-based video tracking algorithm [28].  

For our algorithm, we fixed the number of samples per frame at 80 (cf. region proposal systems commonly rely on thousands of image queries [9]). We report the search summary statistics for IOU (intersection over union) for each model. 
\begin{table}
\begin{center}
\begin{tabular}{|c c c c c|}
\hline
& TM & MOSSE & ADNET & SDBTA (ours) \\
\hline mean IOU & 0.26 & 0.10 & 0.47 & \textbf{0.56}  \\
\hline std IOU & 0.22 & 0.25 & 0.23 & \textbf{0.17} \\
\hline
\end{tabular}
\newline
\end{center}
\caption{Experimental results summary.}
\end{table}

Beyond these strong quantitative tracking results, we additionally observed that the comparison models suffered from either significant long-term tracking deterioration or episodic instability (see Figure 3). The SDBTA algorithm in general did not exhibit this behavior based on our experimental trials.
 \begin{figure}
 \begin{center}
 \includegraphics[scale=0.30]{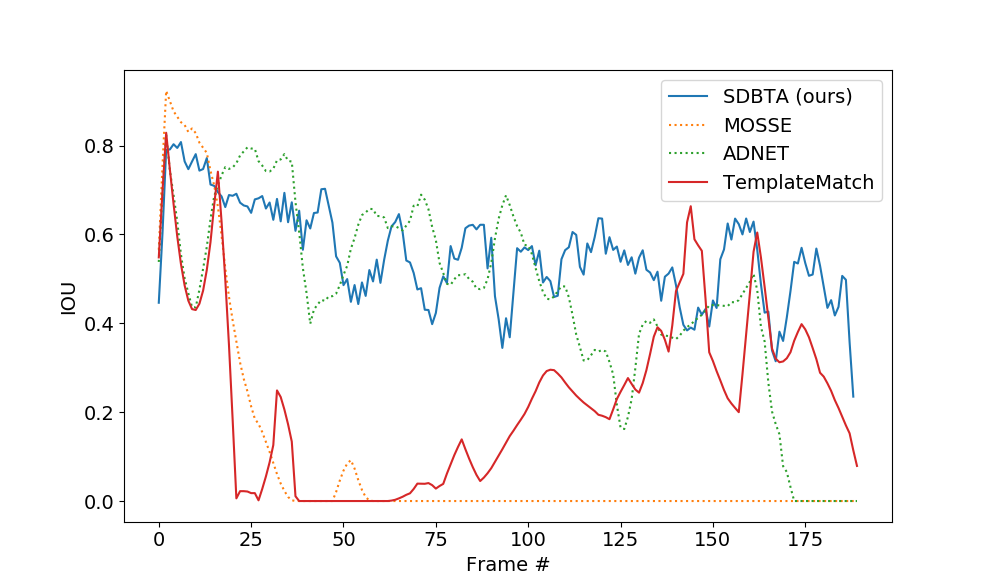}
  \caption{The graph shows the general stability of the SDBTA tracker for a representative test video, 'tc_boat_ce1' (T = 200 frames); IOU is represented by the vertical axis and the frame number corresponds with the horizontal axis. By comparison, the MOSSE tracker essentially fails to track after frame 30; TM fails to track for nearly half of the duration of the video (frames 25-100); and ADNET fails to track after frame 170.}  \end{center}
  \label{fig:GPNIPS}
\end{figure}
\section{Future Work}
 
While the present algorithm has already demonstrated its effectiveness in video tracking, we nevertheless believe it can be further improved in the near future. We intend to expand the current approach to accommodate the following enhancements: (1) GP-enabled multiscaling (so the GP is generated in five dimensions, including space, size and time); (2) adaptive Bayesian optimization (ABO) which adaptively alters the bounds and sample constraints at each frame for optimizing the acquisition function based on the learned time-related length-scale parameter [25]; (3) we anticipate furthermore that incorporating a fully-convolutional [15] architecture into the Siamense conv-net with our current pipeline will yield faster than real-time video tracking with the added benefits of BNP. Following [26] the current research can be augmented to include visual context models for structured image and video types to be used with video scene and behavior recognition; various numerical optimization techniques can further improve the efficiency and speed of our GP-based video tracking, including [27]. We believe that the current research has significant potential for widespread real-world use, including applications to surveillance, high-level scene understanding in computer vision systems, and a myriad of commercial and consumer-based applications. 

\section{References}
\small
[1] Roberto Brunelli. 2009. Template Matching Techniques in Computer Vision: Theory and Practice. Wiley Publishing.  \newline
[2] David G. Lowe. 1999. Object Recognition from Local Scale-Invariant Features. In Proceedings of the International Conference on Computer Vision-Volume 2 - Volume 2 (ICCV '99), Vol. 2. IEEE Computer Society, Washington, DC, USA, 1150-. \newline
[3] George Nebehay and Roman P. Pflugfelder. 2014. Consensus-based Matching and Tracking of Keypoints for Object Tracking. In (IEEE) Winter Conference on Applications of Computer Vision, Steamboat Springs, CO, USA, 862-869. \newline
[4] Avneet Dalal and Bill Triggs. 2005. Histograms of Oriented Gradients for Human Detection. In Proceedings of the 2005 IEEE Computer Society Conference on Computer Vision and Pattern Recognition (CVPR'05) - Volume 1 - Volume 01 (CVPR '05), Vol. 1. IEEE Computer Society, Washington, DC, USA, 886-893. \newline
[5] George Nebehay and Roman P. Pflugfelder. 2015. Clustering of Static-Adaptive Correspondences for Deformable Object Tracking. In (IEEE) Conference on Computer Vision and Pattern Recognition (CVPR), Boston, MA, USA, 2784-2791. \newline
[6] Dorin Comaniciu and Peter Meer. 2002. Mean Shift: A Robust Approach Toward Feature Space Analysis. IEEE Trans. Pattern Anal. Mach. Intell. 24, 5 (May 2002), 603-619. DOI=http://dx.doi.org/10.1109/34.1000236 \newline
[7] B. A. Draper, D. S. Bolme, J. R. Beveridge and Y. M. Lui, "Visual object tracking using adaptive correlation filters," 2010 IEEE Computer Society Conference on Computer Vision and Pattern Recognition(CVPR), San Francisco, CA, USA, 2010, pp. 2544-2550.
doi:10.1109/CVPR.2010.5539960  \newline
[8] "MUlti-Store Tracker (MUSTer): a Cognitive Psychology Inspired Approach to Object Tracking", Zhibin Hong, Zhe Chen, Chaohui Wang, Xue Mei, Danil Prokhorov, and Dacheng Tao, IEEE Conference on Computer Vision and Pattern Recognition (CVPR) 2015, Boston, USA  \newline
[9] Ross Girshick. 2015. Fast R-CNN. In Proceedings of the 2015 IEEE International Conference on Computer Vision (ICCV) (ICCV '15). IEEE Computer Society, Washington, DC, USA, 1440-1448. DOI=http://dx.doi.org/10.1109/ICCV.2015.169 \newline
[10] David Held and Sebastian Thrun and Silvio Savarese. 2017. Learning to Track at 100 FPS with Deep Regression Networks. European Conference on Computer Vision (ECCV). Springer \newline
[11] Zhenhua Fan, Hongbing Ji, and Yongquan Zhang. 2015. Iterative particle filter for visual tracking. Image Commun. 36, C (August 2015), 140-153.  \newline
[12] Shiuh-Ku Weng, Chung-Ming Kuo, and Shu-Kang Tu. 2006. Video object tracking using adaptive Kalman filter. J. Vis. Comun. Image Represent. 17, 6 (December 2006), 1190-1208. \newline
[13] R. Tao and E. Gavves and A. Smeulders. 2016. Siamese Instance Search for Tracking. Computer Vision and Pattern Recognition (CVPR). \newline
[14] Koch, Gregory, Zemel, Richard and Salakhutdinov, Ruslan. "Siamese Neural Networks for One-shot Image Recognition." Paper presented at the meeting of the , 2015. \newline
[15] Bertinetto, Luca et al. “Fully-Convolutional Siamese Networks for Object Tracking.” Computer Vision – ECCV 2016 Workshops (2016): 850–865. Crossref. Web. \newline
[16] Alex Krizhevsky, Ilya Sutskever, and Geoffrey E. Hinton. 2012. ImageNet classification with deep convolutional neural networks. In Proceedings of the 25th International Conference on Neural Information Processing Systems - Volume 1 (NIPS'12), F. Pereira, C. J. C. Burges, L. Bottou, and K. Q. Weinberger (Eds.), Vol. 1. Curran Associates Inc., USA, 1097-1105. \newline
[17] Yann LeCun and Yoshua Bengio. 1998. Convolutional networks for images, speech, and time series. In The handbook of brain theory and neural networks, Michael A. Arbib (Ed.). MIT Press, Cambridge, MA, USA 255-258. \newline
[18] Olga Russakovsky*, Jia Deng*, Hao Su, Jonathan Krause, Sanjeev Satheesh, Sean Ma, Zhiheng Huang, Andrej Karpathy, Aditya Khosla, Michael Bernstein, Alexander C. Berg and Li Fei-Fei. (* = equal contribution) ImageNet Large Scale Visual Recognition Challenge. IJCV, 2015.\newline
[19] Luca Bertinetto and Jack Valmadre and Joao Henriques and Andrea Vedaldi and Phillip Torr. 2016. 
Fully-Convolutional Siamese Networks for Object Tracking. ECCV. \newline
[20] Jurgen Branke. 2001. Evolutionary Optimization in Dynamic Environments. Kluwer Academic Publishers, Norwell, MA, USA. \newline
[21] Carl Edward Rasmussen and Christopher K. I. Williams. 2005. Gaussian Processes for Machine Learning (Adaptive Computation and Machine Learning). The MIT Press. \newline
[22] Tony Jebara, Risi Kondor, and Andrew Howard. 2004. Probability Product Kernels. J. Mach. Learn. Res. 5 (December 2004), 819-844. \newline
[23] Naiyan Wang, Jianping Shi, Dit-Yan Yeung, and Jiaya Jia. 2015. Understanding and Diagnosing Visual Tracking Systems. In Proceedings of the 2015 IEEE International Conference on Computer Vision (ICCV) (ICCV '15). IEEE Computer Society, Washington, DC, USA, 3101-3109. DOI=http://dx.doi.org/10.1109/ICCV.2015.355  \newline
[24] Matej Kristan and Jiri Matas and Alevs Leonardis and Tomas Vjori and Roman Pflugfelder and Gustavo Fernandez and George Nebehay and Fatih Porikli and Luka Vcehovin. 2016. A Novel Performance Evaluation Methodology for Single-Target Trackers. PAMI. \newline
[25]  Favour M. Nyikosa and Michael A. Osbore and Stephen J. Roberts. 2018. Bayesian Optimization for Dynamic PRoblems. arXiv.1803.03432 \newline
[26] Anthony Rhodes and Jordan Witte and Bruno Jedynak and Melanie Mitchell. 2018. Gaussian Processes with Context-Supported Priors for Active Object Localization. International Joint Conference on Neural Networks (IJCNN). \newline
[27] Seth Flaxman and Andrew Wilson and Daniel Neill and Hannes Nickisch and Alex Smola. 2015. Fast Kronecker Inference in Gaussian Processes with non-Gaussian Likelihoods. Proceedings of the 32nd International Conference on Machine Learning. \newline
[28] Sangdoo Yun and Jongwon Choi and Youngjoon Yoo and Kimin Yun and Jin Young Choi. Action-Decision Networks for Visual Tracking with Deep Reinforcement Learning. The IEEE Conference on Computer Vision and Pattern Recognition (CVPR 2017). \newline
[29] Briechle, Kai and Hanebeck, Uwe. (2001). Template matching using fast normalized cross correlation. Proceedings of SPIE - The International Society for Optical Engineering. 4387. 10.1117/12.421129. \newline
[30] Eric Brochu and Vlad M. Cora and Nando de Freitas. 2010. A Tutorial on Bayesian Optimization of Expensive Cost Functions, with Application to Active User Modeling and Hierarchical Reinforcement Learning. arXiv:1012.2599v1 \newline
[31] D. Lizotte. Practical Bayesian Optimization. PhD thesis, University
of Alberta, Edmonton, Alberta, Canada, 2008. \newline
[32] Kristan, Matej et al. 2015. The Visual Object Tracking VOT2014 Challenge Results. ECCV 2014 Workshops. \newline
[33] Matej Kristan et al. 2017. The Visual Object Tracking VOT2016 Challenge Results. ECCV 2017 Workshops. \newline
  
{\small
\bibliographystyle{ieee}
\bibliography{egbib}
}

\end{document}